\documentclass[runningheads]{llncs}
\usepackage{caption}
\usepackage{arydshln}
\usepackage{tabularray}
\usepackage{siunitx}        
\usepackage{etoolbox} 
\usepackage{graphicx}
\usepackage{amsmath}
\usepackage{amssymb}
\usepackage{booktabs}
\usepackage{bbm}
\usepackage{bm}
\usepackage{multirow}
\usepackage{verbatim}
\usepackage[ruled]{algorithm2e}
\usepackage{colortbl}
\usepackage[dvipsnames]{xcolor}
\usepackage[colorlinks,linkcolor=blue,citecolor=blue]{hyperref}
\usepackage{booktabs,subcaption,amsfonts,dcolumn}
\begin{document}
\title{ShapePU: A New PU Learning Framework  Regularized by Global Consistency for Scribble Supervised Cardiac Segmentation}
\titlerunning{ShapePU}
%
\author{Ke Zhang \and Xiahai Zhuang
\thanks{Xiahai Zhuang is corresponding author. This work was funded by the National Natural Science Foundation of China (grant no. 61971142, 62111530195 and 62011540404) and the development fund for Shanghai talents (no. 2020015)}}
\authorrunning{K Zhang \& X Zhuang}
%
\institute{School of Data Science, Fudan University\\
\url{www.sdspeople.fudan.edu.cn/zhuangxiahai}}
%
\maketitle              

\begin{abstract}
Cardiac  segmentation is an essential step for the diagnosis of cardiovascular diseases.
However, pixel-wise dense labeling is both costly and time-consuming.
Scribble, as a form of sparse annotation, is more accessible than full annotations. 
However, it's particularly challenging to train a segmentation network with weak supervision from scribbles.
To tackle this problem, we propose a new scribble-guided method for cardiac segmentation, based on the Positive-Unlabeled (PU) learning framework and global consistency regularization, and termed as \emph{ShapePU}.
To leverage unlabeled pixels via PU learning, we first present an Expectation-Maximization (EM) algorithm to estimate the proportion of each class in the unlabeled pixels.
Given the estimated ratios, we then introduce the marginal probability maximization to identify the classes of unlabeled pixels.
To exploit shape knowledge, we apply cutout operations to training images, and penalize the inconsistent segmentation results.
Evaluated on two open datasets, \emph{i.e}, ACDC and MSCMRseg, our scribble-supervised ShapePU surpassed the fully supervised approach respectively by 1.4\% and 9.8\% in average Dice, and outperformed the state-of-the-art weakly supervised and PU learning methods by large margins.
Our code is available at \url{https://github.com/BWGZK/ShapePU}. 
\end{abstract}

\begin{figure}[t]
    \centering
    \includegraphics[width=0.99\textwidth]{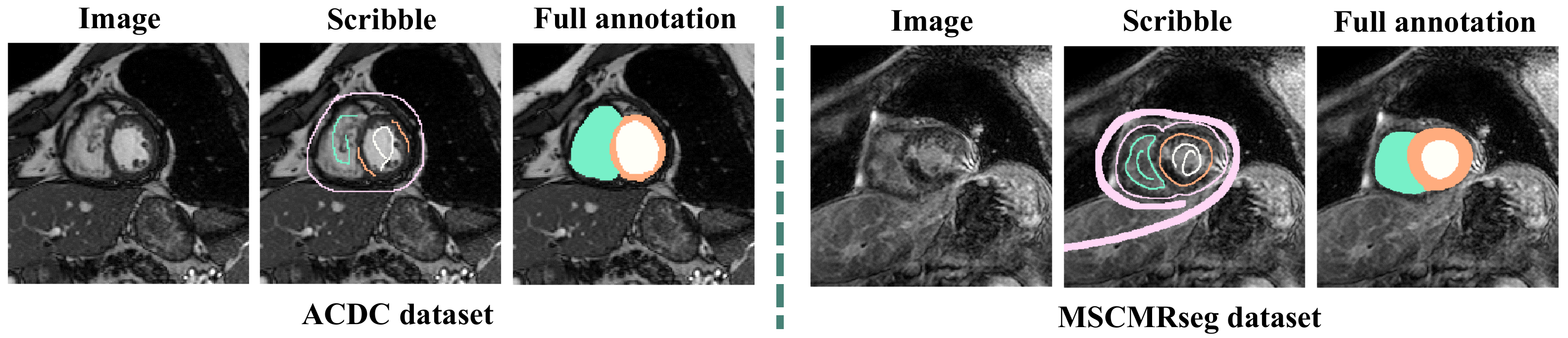}\\[-1ex] 
    \caption{Examples from ACDC and MSCMRseg datasets.}
    \label{fig:samples}
\end{figure}

\section{Introduction}
Curating a large scale of fully annotated dataset is burdensome, particularly in the field of medical image analysis.
However, most of advanced segmentation models are fully supervised and rely on pixel-wise dense labeling~\cite{ronneberger2015u,zhuang2018multivariate,ZHUANG201677}.
To alleviate it, existing literature~\cite{Can2018LearningTS,khoreva2017simple,shi2021marginal,9389796,zhou2019prior} have explored weaker form of annotations (e.g. image-level label, sparse label, noisy label),
among which scribbles in Fig.~\ref{fig:samples} are particularly attractive in medical image segmentation~\cite{Can2018LearningTS}. 
Therefore, we propose to utilize only scribble-annotated data for model training, which is a specific form of weakly supervised segmentation.

Two challenges are presented for scribble-guided segmentation, \textit{i.e.,} insufficient supervision and incomplete shape of the annotated object.
Existing methods exploited  labeled pixels~\cite{bai2018recurrent,ji2019scribble,l2016inscribblesup}, but the supervision from unlabeled pixels is rarely explored.
To capture complete shape features, several methods proposed to learn from unpaired fully annotated segmentation masks, meaning additional resources are required~\cite{Larrazabal2020PostDAEAP,9389796,zhang2020accl}.

To exploit supervision from unlabeled data, a line of researches have been proposed to learn from positive and unlabeled data, well known as \emph{PU learning}~\cite{de1999positive,garg2021mixture,NIPS2017_7cce53cf,letouzey2000learning}. 
This framework is designed for binary classification task and aims to extract negative samples from unlabeled data.
In medical imaging, PU learning has been applied to classification~\cite{nagaya2021embryo} and object detection tasks~\cite{zuluaga2011learning}.
Many methods have been proposed for binary mixture proportion estimate~\cite{bekker2018estimating,garg2021mixture,ramaswamy2016mixture} and PU learning~\cite{du2015convex,du2014analysis,NIPS2017_7cce53cf}.
However, these methods generate independent estimate of mixture ratio for each class, which is unreasonable in multi-class image segmentation.
Correspondingly, \emph{existing PU learning methods for classification task cannot be directly adapted for image segmentation}.

\begin{figure}[t]
    \centering
    \includegraphics[width=0.9\textwidth]{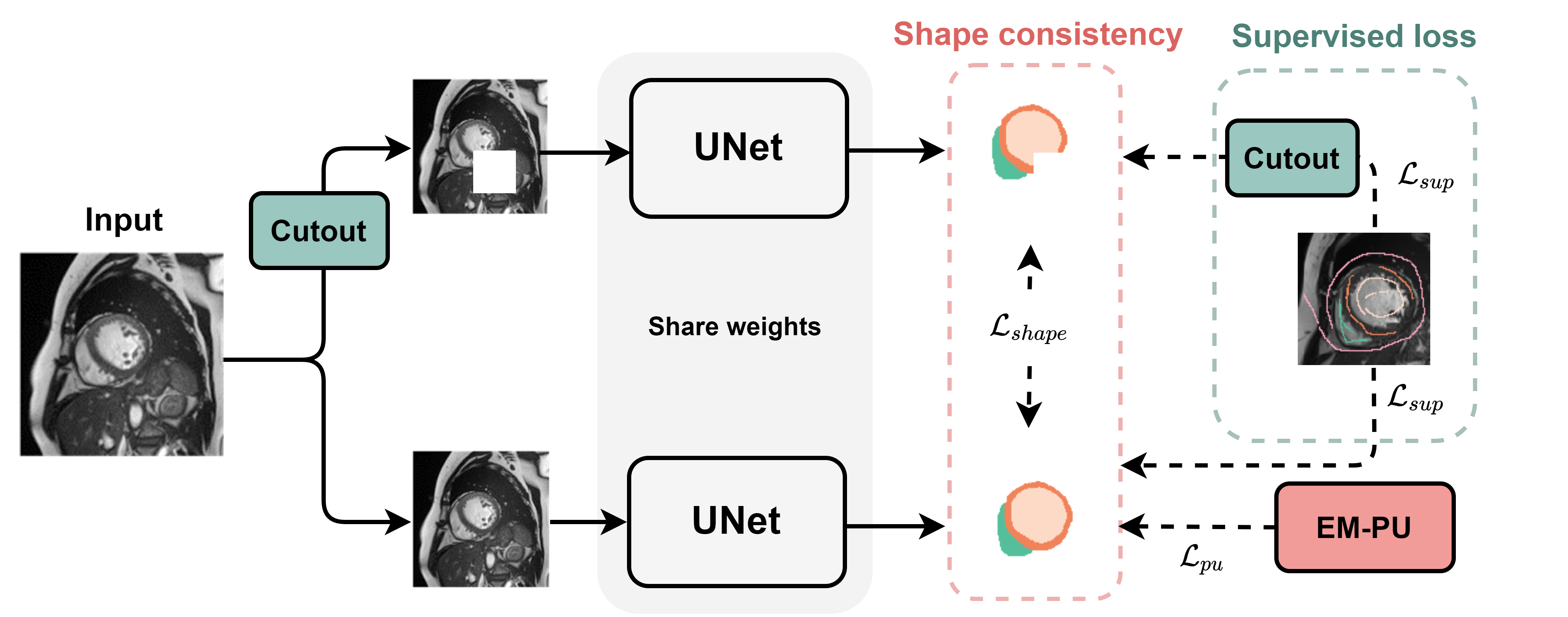}
    \caption{Overview of the proposed ShapePU framework for cardiac segmentation from scribble supervision.}
    \label{fig:pipeline}
\end{figure}

To tackle the above challenges, we propose a novel shape-constrained PU learning method, \emph{i.e.}, \emph{ShapePU}, for scribble-guided cardiac segmentation.
As Fig.~\ref{fig:pipeline} shows, ShapePU consists of a PU learning framework for seeking supervision from unlabeled pixels and consistency constraint for shape regularization.
Firstly, We adopt EM algorithm to estimate the multi-class mixture ratios in unlabeled pixels.
Then, we conduct PU learning to identify classes of unlabeled pixels by maximizing marginal probability.
Finally, we regularize the segmentation by global consistency, which captures shape features by leveraging cutout-equivalence of image.
As illustrated in Fig.~\ref{fig:pipeline}, cutout-equivalence requires the prediction of an image should obtain the same cutout of the input. 
Therefore, ShapePU enables the model to exploit supervision from unlabeled pixels and capture the global shape features.

Our contributions are summarized as follows:
(1) To the best of our knowledge, this is the first PU learning framework formulated for weakly supervised segmentation, incorporated with shape knowledge.
(2) We propose the novel PU learning framework for multi-class segmentation, which implements EM estimation of mixture ratios and PU learning to identify classes of unlabeled pixels.
(3) We introduce the global consistency regularization by leveraging the cutout equivalence, which can enhance the ability of model to capture global shape features. 
(4) The proposed \emph{ShapePU} consistently outperformed the fully supervised methods, the state-of-the-art weakly supervised approaches, and the competitive PU learning methods on two cardiac segmentation tasks. 

\section{Method}

As shown in Fig.~\ref{fig:pipeline}, our method takes UNet~\cite{baumgartner2017exploration} as backbone.
Besides the supervised loss of annotated pixels, we
leverage the unlabeled pixels via a new PU framework regularized by global consistency.
The proposed ShapePU consists of three components: (1) multi-class mixture proportion estimation based on EM algorithm; (2) PU learning step to distinguish unlabeled pixels by maximizing the marginal probability; and (3) the global consistency leveraging cutout equivalence.

\subsection{EM Estimation of Multi-class Mixture Proportion}
For multi-class segmentation with $m+1$ label classes including background $c_0$,
we denote $P_j, j= 0,\cdots,m$ as the class-conditional distributions of label class $c_j, j=0,\cdots,m$, and $p_j$ as its density. 
Let $P_u$ be the distribution of unlabeled pixels with density $p_u$.
We formulate $P_u$ as the mixture of $\{P_j\}_{j=0}^{m}$, \emph{i.e}, $P_u=\sum_{j=0}^{m}\alpha_j P_j$.
$\alpha_j\in[0,1]$ is the mixture proportion of class $c_j$, which satisfying $\sum_{j=0}^m \alpha_j =1$.
In weakly supervised segmentation with scribble annotations, we treat each pixel of label $c_j$ as an i.i.d sample from the class-conditional distribution $P_j$.
Similarly, the unlabeled pixels are taken as i.i.d samples from mixed distribution $P_u$. 
The goal of mixture proportion estimation is to estimate $\{\alpha_j\}_{j=0}^m$.

We employ the EM algorithm~\cite{latinne2001adjusting} to estimate the multi-class mixture proportions. For class $c_j$ and labeled pixel $x$, one has
$\hat{p}_l(x|c_j)=\frac{\hat{p}_l(c_j|x)p(x)}{\hat{p}_l(c_j)}$ based on Bayes' theorem, where $\hat{p}_l(x|c_j)$ denotes the within-class probability of labeled pixels. 
Similarly, we obtain the within-class probability of unlabeled pixels by $\hat{p}_u(x|c_j) = \frac{\hat{p}_u(c_j|x)\hat{p}_u(x)}{\hat{p}_u(c_j)}$.
We assume the within-class probability $\hat{p}(x|c_j)$ of labeled and unlabeled pixels be the same, \emph{i.e.}, $\hat{p}_l(x|c_j)=\hat{p}_u(x|c_j)$. Given the condition that $\sum_{j=0}^m\hat{p}_u(c_j|x)=1$, we can solve that:
\begin{equation}
    \alpha_j = \frac{1}{n_u}\sum_{i=1}^{n_u} \hat{p}_u(c_j |x_i) = \frac{1}{n_u}\sum_{i=1}^{n_u} \frac{\hat{p}_u(c_j)\hat{p}_l(c_j|x_i)/\hat{p}_l(c_j)}{\sum_{j=0}^m[\hat{p}_u(c_j)\hat{p}_l(c_j|x_i)/\hat{p}_l(c_j)]},
    \label{eq1}
\end{equation}
where the mixture ratio $\alpha_j$ equals $\hat{p}_u(c_j)$; $\hat{p}_u(c_j|x_i)$ is the probability of pixel $x_i$ belonging to $c_j$, which is predicted by segmentation network; $\hat{p}_l(c_j)$ is the proportion of class $c_j$ in labeled pixels; $n_u$ is the number of unlabeled pixels.
The mixture ratios $\hat{p}_u(c_j)$ of unlabeled pixels are initialized with the class frequencies of labeled pixels, \textit{i.e.}, $\hat{p}_l(c_j)$.
Then, we substitute the estimated $\alpha_j$ into $\hat{p}_u(c_j)$ on the right side of the formula, and repeat it until the value of $\alpha_j$ converges. 
The detailed derivation of Eq.(\ref{eq1}) is provided in the supplementary material.

\subsection{PU Learning with Marginal Probability}
Given the estimated mixture proportions, we aim to identify the classes of unlabeled pixels.
Firstly, positive samples of each class are distinguished from unlabeled pixels.
Secondly, we design the negative loss ($\mathcal{L}^{-}$) to maximize the marginal probability of negative samples.

For class $c_j$, we rank the unlabeled pixels according to their probability belongs to class $c_j$, which is predicted by the segmentation network. 
After that, the pixels within the $\alpha_j$ proportion are considered as positive samples, denoted as $\Omega_j$. 
The remain $1-\alpha_j$ proportion is taken as a set of negative samples, which is represented as $\bar{\Omega}_j$.

We apply loss only to foreground classes.
Given the observation that the ratios of foreground classes tend to be over-estimated~\cite{garg2021mixture}, we do not directly compute the loss on the predicted positive samples. 
Instead, we apply a loss function to the set of predicted negative samples ($\bar{\Omega}_j$).
Firstly, we fuse other classes together except $c_j$, and denote the fused class as $\bar{c}_j$.
Then the marginal probability of $\bar{c}_j$ is equal to the sum of the probabilities belonging to related classes, \emph{i.e.}, $\hat{p}(\bar{c}_j|x) = \sum_{k=1}^m [\mathbbm{1}_{[k\neq j]}\hat{p}(c_k|x)]$.
Finally, we formulate the negative loss $\mathcal{L}^{-}$ to maximize the marginal probabilities, \emph{i.e.},
\begin{equation}
    \mathcal{L}^{-} =-\sum_{j=1}^m \sum_{i\in \bar{\Omega}_j}\log(\hat{p}\left(\bar{c}_j|x_i)\right).
\end{equation}

\subsection{Global Consistency}
In segmentation tasks, We need to consider not only individual pixels, but also global shape features.
Taking an input image, denoted as $X$, we first randomly \emph{cutout} a square area of $X$.
Let $z$ be the binary cutout mask in $[0,1]$ and $T(\cdot)$ be the transformation of rotation and flipping.
The perturbed image is represented as $X'=T( z \odot X)$.
Then, we require the segmentation result of the image $X$ and masked image $X'$ to be consistent except the cutout area.
Therefore, we have $T(z \odot f(X)) = f(T(z\odot X))$, where $f$ denotes the segmentation network.
Defining $f'(X) = T(z \odot f(X))$, we formulate global consistency loss $\mathcal{L}_{global}$ as:
\begin{equation}
    \mathcal{L}_{global} = \frac{1}{2}\mathcal{L}_{cos}(f'(X), f(X') + \frac{1}{2}\mathcal{L}_{cos}(f(X'), f'(X)),
\end{equation}
where $\mathcal{L}_{cos}$ indicates cosine similarity distance, \emph{i.e.}, $\mathcal{L}_{cos}(a,b) = -\frac{a\cdot b}{(||a||_2\cdot||b||_2)}$.

Given scribble annotations, we calculate the supervised cross entropy loss $\mathcal{L}^{+}$ for annotated pixels of both $X$ and $X'$.  
Let the set of labeled pixels be $\Omega_l$ and the label of pixel $x_i$ be a vector of dimension $m$, \emph{i.e.}, $\bm{y}_i=\{y_{i1},\cdots, y_{im}\}$.
We denote the predicted label of pixel $x_i$ as $\hat{\bm{y}_i} = \{\hat{y}_{i1},\cdots, \hat{y}_{im}\}$.
Then, the supervised cross entropy loss $\mathcal{L}^{+}$ is written as:
\begin{equation}
    \mathcal{L}^{+}=-\sum_{i\in\Omega_l}\sum_{j=1}^m \left[y_{ij}\log(\hat{y}_{ij})\right].
\end{equation}
Finally, the optimization objective is represented as:
\begin{equation}
    \mathcal{L} =  \mathcal{L}^{+} + \lambda_1 \mathcal{L}^{-} + \lambda_2 \mathcal{L}_{global},
\end{equation}
where $\lambda_1$ and $\lambda_2$ are balancing parameters.

\section{Experiment}
In the experiments, we first performed the ablation study on the ACDC dataset~\cite{8360453}.
Then, we compared our ShapePU to weakly supervised methods and PU learning approaches using ACDC and MSCMRseg dataset~\cite{zhuang2016multivariate,zhuang2018multivariate}, respectively.
We further analyzed the stability of model training and presented estimated mixture ratios in the supplementary material.
\subsection{Experiment Setup}
\textbf{Datasets.} 
\textbf{ACDC}\footnote{\url{https://www.creatis.insa-lyon.fr/Challenge/acdc/databasesTraining.html}} consists of fully annotated cardiac MR images from 100 patients.
The goal is to automatically segment right ventricle (RV), left ventricle (LV) and myocardium (MYO). 
We randomly divided the 100 subjects into 3 sets of 70 (training), 15 (validation), and 15 (test) subjects for experiments. 
The expert-made scribble annotations in \cite{9389796} were leveraged for weak-supervision studies.
\textbf{MSCMRseg}\footnote{\url{http://www.sdspeople.fudan.edu.cn/zhuangxiahai/0/mscmrseg19/data.html}} includes late gadolinium enhancement (LGE) cardiac MR images from 45 patients with cardiomyopathy. 
MSCMRseg is more challenging compared to ACDC, as LGE CMR segmentation per se is more complex and the training set is smaller. 
We generated scribble annotations for RV, LV and MYO following the similar protocol of \cite{9389796}. 
We randomly divided the 45 images into 25 training images, 5 validation images, and 15 test images.
\newline \textbf{Implementation.} We warmly started training the networks with supervised loss and global consistency loss for 100 epochs, and then invoked the negative loss. 
Since the images are of different resolutions, we first re-sampled all images to a fixed resolution of $1.37\times1.37$mm and then extracted the central patch of size $212\times212$ as input.
Then, we normalized the intensity to zero mean and unit variance.
A square area of $32\times 32$ was randomly cut out for each image.
Hyper-parameters $\lambda_1$ and $\lambda_2$ were empirically set to be $1$ and $0.05$.
All models were trained with batch size 16 and learning rate 1e$^{-4}$.
For testing, we kept the largest connected area of foreground to eliminate false positives.
Dice scores and Haussdorff Distance (HD) are reported to measure the accuracy of segmentation models.
We implemented our framework with Pytorch and conducted experiments on one NVIDIA 3090Ti 24GB GPU for 1000 epochs.

\subsection{Ablation study}
\begin{table}[t]
	\caption{Results in Dice scores of the ablation study. \textbf{Bold} denotes the best result, \underline{underline} indicates the  best but one. Significant improvement compared to previous model given by Wilcoxon test (p$<$0.05) is denoted with $^\dag$.}\label{tab1}
	\begin{center}
		\resizebox{\textwidth}{!}{
			\begin{tabular}{ccccccccc}
			\toprule
				Methods& $\mathcal{L}^{+}$ & Cutout & $\mathcal{L}^{-}$ & $\mathcal{L}_{global}$ & LV & MYO & RV &Avg\\
				\midrule
				\#1&$\checkmark$&$\times$&$\times$&$\times$&.808$\pm$.161&.749$\pm$.099&.779$\pm$.133&\multicolumn{1}{c}{.779}\\
				\#2&$\checkmark$&$\checkmark$&$\times$&$\times$&.815$\pm$.172&.758$\pm$.134&.817$\pm$.123$^\dag$&\multicolumn{1}{c}{.797$^\dag$}\\
				\textbf{\#3}&$\checkmark$&$\times$&$\checkmark$&$\times$&\underline{.870$\pm$.141$^\dag$}&\underline{.798$\pm$.104$^\dag$}&.832$\pm$.133&\multicolumn{1}{c}{.833$^\dag$}\\
			\#4&$\checkmark$&$\checkmark$&$\checkmark$&$\times$&.859$\pm$.150&.794$\pm$.113&\underline{.850$\pm$.104}&\multicolumn{1}{|c}{\underline{.834}}\\	\textbf{ShapePU}&$\checkmark$&$\checkmark$&$\checkmark$&$\checkmark$&\textbf{.888$\pm$.103$^\dag$}&\textbf{.813$\pm$.095$^\dag$}&\textbf{.854$\pm$.089}&\multicolumn{1}{c}{\textbf{.851}$^\dag$}\\
				\bottomrule
		\end{tabular}}
	\end{center}
\end{table}

We performed an ablation study to verify effects of the two key components of the proposed  ShapePU, \emph{i.e.}, the negative loss ($\mathcal{L}^{-}$) and the global consistency loss ($\mathcal{L}_{global}$).
Table~\ref{tab1} reports the results.  
One can see that cutout augmentation showed marginal improvement over model \#1.
Having supervision for unlabeled pixels, model \#3 included the negative loss ($\mathcal{L}^{-}$) and obtained remarkable performance gain over model \#1 by 5.5\% ($83.3\%$ vs $77.8\%$) on the average Dice.
When combined with Cutout, PU (without shape) improved the average Dice marginally from $83.3\%$ to $83.4\%$. Cutout enhances the localization ability, but may change the shape of target structure. Therefore, it could be difficult for the segmentation model to learn the shape priors, leading to the performance drop in some structures. When combined with global consistency ($\mathcal{L}_{global}$), which overcomes the disadvantage by requiring the cutout equivalence, the performance is evidently better with 
a significant improvement of 1.7\% on average Dice ($85.1\%$ vs $83.4\%$).
This indicated that the combination of PU learning and global consistency endows the algorithm with supervision of unlabeled pixels and with knowledge of global shapes.

\begin{table}[t]
\caption{Comparisons of ShapePU trained on 35 scribbles with other scribble-guided models, GAN-based models, and fully supervised method. The results cited from \cite{9389796}, which did not report standard deviation, are denoted with *.}
\label{tab2}
 	\begin{center}
\resizebox{0.9\textwidth}{!}{
		\begin{tabular}{ccccccccc}
				\toprule
				\multirow{2}{*}{Methods}& \multicolumn{4}{c}{Dice} &\multicolumn{4}{c}{HD (mm)}\\
				\cmidrule(lr){2-5}\cmidrule(lr){6-9}
				&LV & MYO & RV &Avg& LV & MYO & RV &Avg\\
				\midrule\midrule
				\multicolumn{9}{l}{35 scribbles}\\
				\midrule
				\multicolumn{1}{c|}{PCE}&.795$\pm$.193&.745$\pm$.143&\underline{.755$\pm$.204}&\multicolumn{1}{c|}{.765}&16.9$\pm$24.1&\underline{23.4$\pm$25.7}&40.6$\pm$28.4&26.9\\
				\multicolumn{1}{c|}{WPCE*}&.784&.675&.563&\multicolumn{1}{c|}{.674}&97.4&99.6&120.5&105.8\\
				\multicolumn{1}{c|}{CRF*}&.766&.661&.590&\multicolumn{1}{c|}{.672}&99.6&103.2&117.8&106.9\\
				\multicolumn{1}{c|}{\textbf{ShapePU (ours)}}&.860$\pm$.122&.791$\pm$.091&\textbf{.852$\pm$.102}&\multicolumn{1}{c|}{\textbf{.834}}&\textbf{12.0$\pm$10.5}&\textbf{13.8$\pm$10.5}&\textbf{11.9$\pm$7.60}&\textbf{12.6}\\
			    \midrule\midrule
			    \multicolumn{9}{l}{35 scribbles + 35 unpaired full annotations}\\
			    \midrule
				\multicolumn{1}{c|}{PostDAE*}&.806&.667&.556&\multicolumn{1}{c|}{.676}&80.6&88.7&103.4&90.9\\
				\multicolumn{1}{c|}{ACCL*}&.878&.797&.735&\multicolumn{1}{c|}{.803}&\underline{16.6}&28.8&26.1&\underline{23.8}\\
			    \multicolumn{1}{c|}{MAAG*}&\textbf{.879}&\textbf{.817}&.752&\multicolumn{1}{c|}{\underline{.816}}&25.2&26.8&\underline{22.7}&24.9\\
			    \midrule\midrule
			    \multicolumn{9}{l}{35 full annotations}\\
			    \midrule
			    \multicolumn{1}{c|}{UNet$_\text{F}$}&.849$\pm$.152&.792$\pm$.140&.817$\pm$.151&\multicolumn{1}{c|}{.820}&15.7$\pm$13.9&13.8$\pm$12.2&13.2$\pm$13.4&14.2\\
				\bottomrule
		\end{tabular} }
 	\end{center}
 \end{table}	

\begin{figure}[t]
    \centering
    \includegraphics[width=0.85\textwidth]{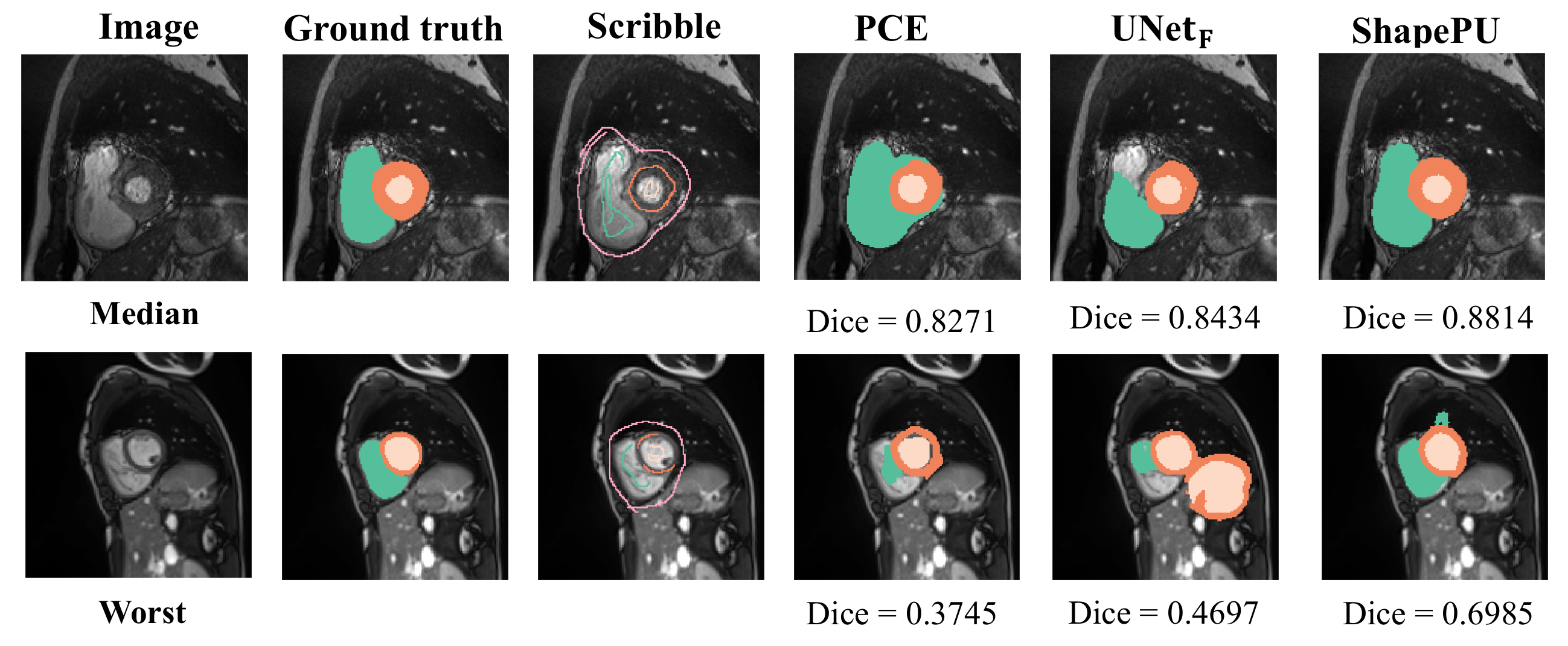}
    \caption{Visualization on two typical ACDC cases for illustration and comparison.}
    \label{fig:fig1}
\end{figure}

\subsection{Comparisons with weakly supervised methods}
We performed two groups of experiments.
One included three scribble-guided models, \emph{i.e.}, partial cross entropy (PCE)~\cite{tang2018normalized}, weighted partial cross entropy (WPCE)~\cite{9389796}, conditional random fields post-processing (CRF)~\cite{zheng2015conditional}.
The other consisted of three GAN-based models trained with \emph{additional unpaired full annotations} to provide shape priors, \emph{i.e.}, post-processing with denoising auto-encoders (PostDAE)~\cite{Larrazabal2020PostDAEAP}, adversarial constrained CNN (ACCL)~\cite{zhang2020accl}, multi-scale attention gates (MAAG)~\cite{9389796}.
Finally, the results from \emph{fully supervised} UNet~\cite{baumgartner2017exploration} (UNet$_\text{F}$), were provided for reference.

Table~\ref{tab2} provides the quantitative results.
ShapePU outperformed all other methods by a large margin.
Notably, the weakly supervised ShapePU matched the performance of fully supervised UNet (UNet$_\text{F}$) with a significant better HD on LV (p=0.041), demonstrating its outstanding learning ability from scribble supervision.
Fig.~\ref{fig:fig1} visualizes results from the median and worst cases selected using the mean Dice of compared methods.
ShapePU was more robust to diverse shapes and densities of heart than the fully supervised UNet (UNet$_\text{F}$), thanks to the effective learning of unlabeled pixels and shape features.
\begin{table}[t]
	\caption{The performance on MSCMRseg and comparisons with other PU methods based on scribble supervision. GT$\alpha$ indicates the ground truth mixture proportions of $\alpha$ are provided for model training; UNet$_\text{F}$ is a fully supervised approach solely for reference, and n/a means not applicable.}\label{tab3}
	\begin{center}
		\resizebox{\textwidth}{!}{
			\begin{tabular}{cccccccccc}
				\toprule
				\multirow{2}{*}{Methods}&\multirow{2}{*}{GT$\alpha$}&\multicolumn{4}{c}{Dice}&\multicolumn{4}{c}{HD (mm)}\\
				\cmidrule(lr){3-6}\cmidrule(lr){7-10}
				 &  &LV & MYO & RV &\multicolumn{1}{c|}{Avg}&LV & MYO & RV &Avg\\
				\midrule
				\multicolumn{1}{c|}{PCE}&\multicolumn{1}{c|}{$\times$}&.514$\pm$.078&.582$\pm$.067&.058$\pm$.023&\multicolumn{1}{c|}{.385}&259.4$\pm$14.2&228.1$\pm$21.4&257.4$\pm$12.4&248.3\\
				\multicolumn{1}{c|}{(TED)$^n$}&\multicolumn{1}{c|}{$\times$}&.524$\pm$.098&.443$\pm$.122&.363$\pm$.125&\multicolumn{1}{c|}{.443}&107.8$\pm$63.6&82.0$\pm$70.2&30.0$\pm$21.3&73.2\\
				\multicolumn{1}{c|}{CVIR}&	\multicolumn{1}{c|}{$\checkmark$}&519$\pm$.042&.519$\pm$.081&.443$\pm$.089&\multicolumn{1}{c|}{.493}&73.2$\pm$7.22&65.3$\pm$68.9&69.3$\pm$86.3&69.3\\
				\multicolumn{1}{c|}{nnPU}&\multicolumn{1}{c|}{$\checkmark$}&.516$\pm$.075&.536$\pm$.085&.442$\pm$.121&\multicolumn{1}{c|}{.498}&76.7$\pm$11.0&41.7$\pm$13.3&\textbf{23.9$\pm$20.6}&47.4\\
				\multicolumn{1}{c|}{\textbf{PU w/o Shape(ours)}}&\multicolumn{1}{c|}{$\times$}&\underline{.911$\pm$.042}&\underline{.808$\pm$.063}&\underline{.793$\pm$.101}&\multicolumn{1}{c|}{\underline{.837}}&\underline{45.4$\pm$73.8}&\underline{36.5$\pm$61.1}&43.1$\pm$52.5&\underline{41.7}\\
				\multicolumn{1}{c|}{\textbf{ShapePU(ours)}}&	\multicolumn{1}{c|}{$\times$}&\textbf{.919$\pm$.029}&\textbf{832$\pm$.042}&\textbf{.804$\pm$.123}&\multicolumn{1}{c|}{\textbf{.852}}&\textbf{10.3$\pm$13.0}&\textbf{10.6$\pm$10.1}&\underline{24.3$\pm$19.6}&\textbf{15.0}\\
				\midrule
				\midrule
				\multicolumn{1}{c|}{UNet$_\text{F}$}&	\multicolumn{1}{c|}{n/a}&.856$\pm$.040&.722$\pm$.061 &.684$\pm$.122&\multicolumn{1}{c|}{.754}&17.2$\pm$20.6&16.6$\pm$15.4&81.5$\pm$7.06&38.4\\
				\bottomrule
		\end{tabular}
		}
	\end{center}
\end{table}	

\begin{figure}[t]
    \centering
    \includegraphics[width=\textwidth]{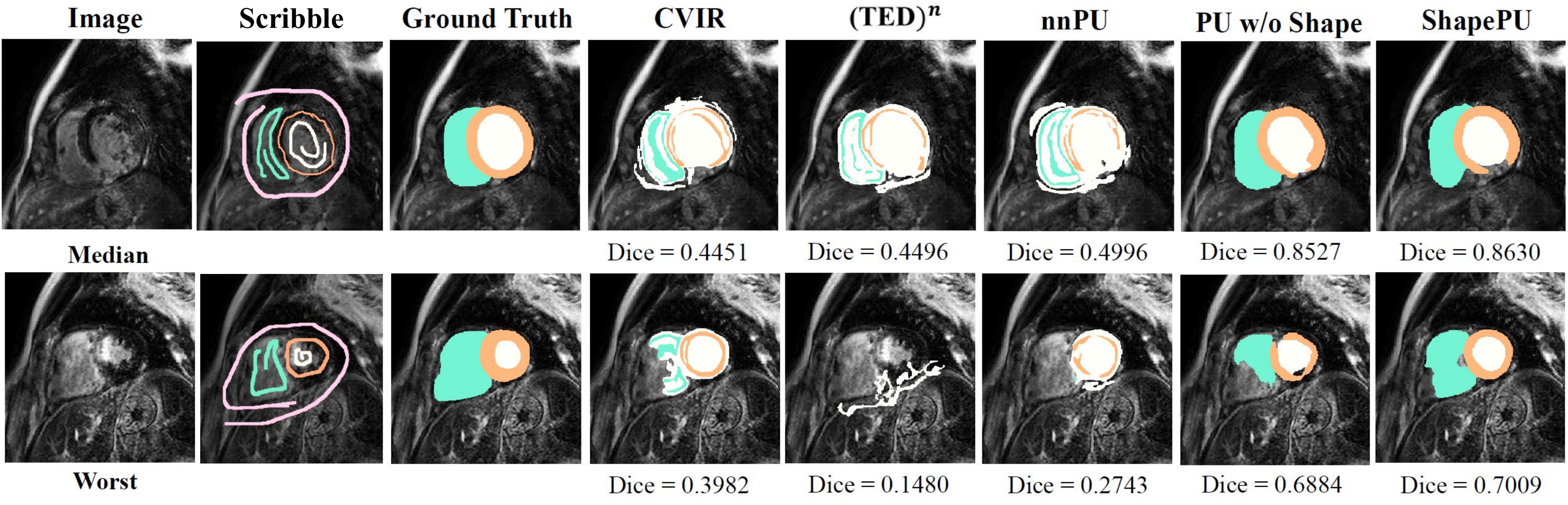}
    \caption{Visualization on two typical MSCMRseg cases for illustration and comparisons with PU learning methods.}
    \label{fig:fig2}
\end{figure}
\subsection{Comparisons with PU learning methods}
We performed the scribble-guided segmentation on MSCMR LGE CMR images and  compared ShapePU with other state-of-the-art PU learning approaches, including Transform-Estimate-Discard ((TED)$^n$)~\cite{garg2021mixture}, PU learning with conditional value ignoring risk (CVIR)~\cite{garg2021mixture}, and positive-unlabeled learning with non-negative risk estimator (nnPU)~\cite{NIPS2017_7cce53cf}.
 
Table~\ref{tab3} reports the quantitative results. 
Without additional information of ground truth mixture ratio $\alpha$ (GT $\alpha$), our basic PU learning framework (PU w/o Shape) still significantly outperformed the peers with at least $33.9\%$ average Dice, demonstrating the challenge of directly applying PU learning for image segmentation, and the effectiveness of our PU formulation for this task.
By leveraging shape features, the proposed ShapePU further boosted the performance to $85.2\%$ (85.2\% vs 83.7\%), with an improvement of $26.7$mm (15.0mm vs 41.7mm) on average HD.
Fig.~\ref{fig:fig2} visualizes the worst and median cases selected using the average Dice scores of all compared PU methods.
One can observe that ShapePU achieved the best performance on both the two cases.

\section{Conclusion}
This paper presents a shape-constrained PU learning (ShapePU) method for weakly supervised cardiac segmentation.
To provide supervision for unlabeled pixels, we adopted EM estimation for the mixture ratios in unlabeled pixels, and employed PU framework to distinguish the classes of unlabeled pixels.
To tackle incomplete shape of scribbles, we proposed the shape-consistency loss to regularize cutout equivalence and capture global shape of the heart. The proposed ShapePU has been evaluated on two publicly available datasets, and achieved the new state-of-the-art performance.

%
%
%
\bibliographystyle{splncs04}
\bibliography{ref}
\end{document}